\documentclass[10pt,twocolumn,letterpaper]{article}

\usepackage{cvpr}              
\usepackage[utf8]{inputenc} 
\usepackage[T1]{fontenc}    
\usepackage{url}           
\usepackage{amsfonts}       
\usepackage{nicefrac}      
\usepackage{microtype}      
\usepackage{xcolor}        
\usepackage{color}
\usepackage{arydshln}
\usepackage{amsmath}
\usepackage{multirow}
\usepackage{graphicx}   
\usepackage{array} 
\usepackage{amsmath}
\usepackage{adjustbox}
\usepackage{float}
\usepackage{makecell} 
\usepackage{enumitem}
\usepackage{siunitx} 
\usepackage{afterpage}
\usepackage{tabularx}
\usepackage{caption}  
\usepackage{subcaption}
\usepackage{graphicx}
\usepackage{amsmath}
\usepackage{amssymb}
\usepackage{booktabs}
\usepackage{CJKutf8}
\usepackage{overpic}
\usepackage{wrapfig}
\usepackage{enumitem}
\usepackage{diagbox}
\usepackage{float}
\usepackage[table]{xcolor}

\definecolor{cvprblue}{rgb}{0.21,0.49,0.74}
\usepackage[pagebackref,breaklinks,colorlinks,allcolors=cvprblue]{hyperref}

\title{OrdMoE: Preference Alignment via Hierarchical Expert Group Ranking in Multimodal Mixture-of-Experts LLMs}

\author{
  Yuting Gao$^{1}$\thanks{The first two authors contributed equally.}
  \quad Weihao Chen$^{1*}$
  \quad Lan Wang$^{1}$
  \quad Ruihan Xu $^{1,2}$
  \quad Qingpei Guo$^{1}$ \\[0.25cm]
  \small
  $^1$AntGroup \quad 
  $^2$Peking University \\ [0.25cm]
  \tt\small yutinggao.sh@gmail.com
}

\begin{document}
\maketitle
\begin{abstract} 
Preference learning has recently emerged as a pivotal strategy for post-training alignment of Multimodal Large Language Models (MLLMs). However, existing approaches predominantly rely on external human-annotated preference data, which is costly and labor-intensive to collect. In this work, we propose OrdMoE, a novel preference alignment framework that bypasses the reliance on external human preferences entirely by leveraging intrinsic signals within Mixture-of-Experts (MoE) architectures. Specifically, we observe that the router’s expert selection scores implicitly encode a quality-aware ranking of responses (\textit{i.e.} higher-scoring experts consistently generate higher-quality outputs). Building on this insight, OrdMoE constructs an internal preference hierarchy by grouping experts into ranked tiers based on their per-token routing scores and activating each tier separately to produce a sequence of responses with increasing quality. This yields a zero-cost, self-supervised preference ordering over generated responses, which can be directly optimized using standard preference learning objectives. Extensive experiments across multiple multimodal benchmarks demonstrate that OrdMoE significantly enhances both alignment and overall performance of multimodal Mixture-of-Experts LLMs, achieving competitive results without requiring any human-annotated preference data.

\end{abstract}    
\section{Introduction}
\label{sec:Intro}
Preference learning has become a standard practice for aligning large language models with desired behaviors, significantly enhancing their general capabilities through feedback-driven optimization. In the multimodal setting, state-of-the-art Multimodal Large Language Models (MLLMs), including Gemini~2.5~\cite{comanici2025gemini} and Qwen2.5-VL~\cite{bai2025qwen2}, routinely adopt preference-based fine-tuning after supervised training to improve overall performance across diverse tasks.
\begin{figure}[t]
    \centering
    \includegraphics[width=\linewidth]{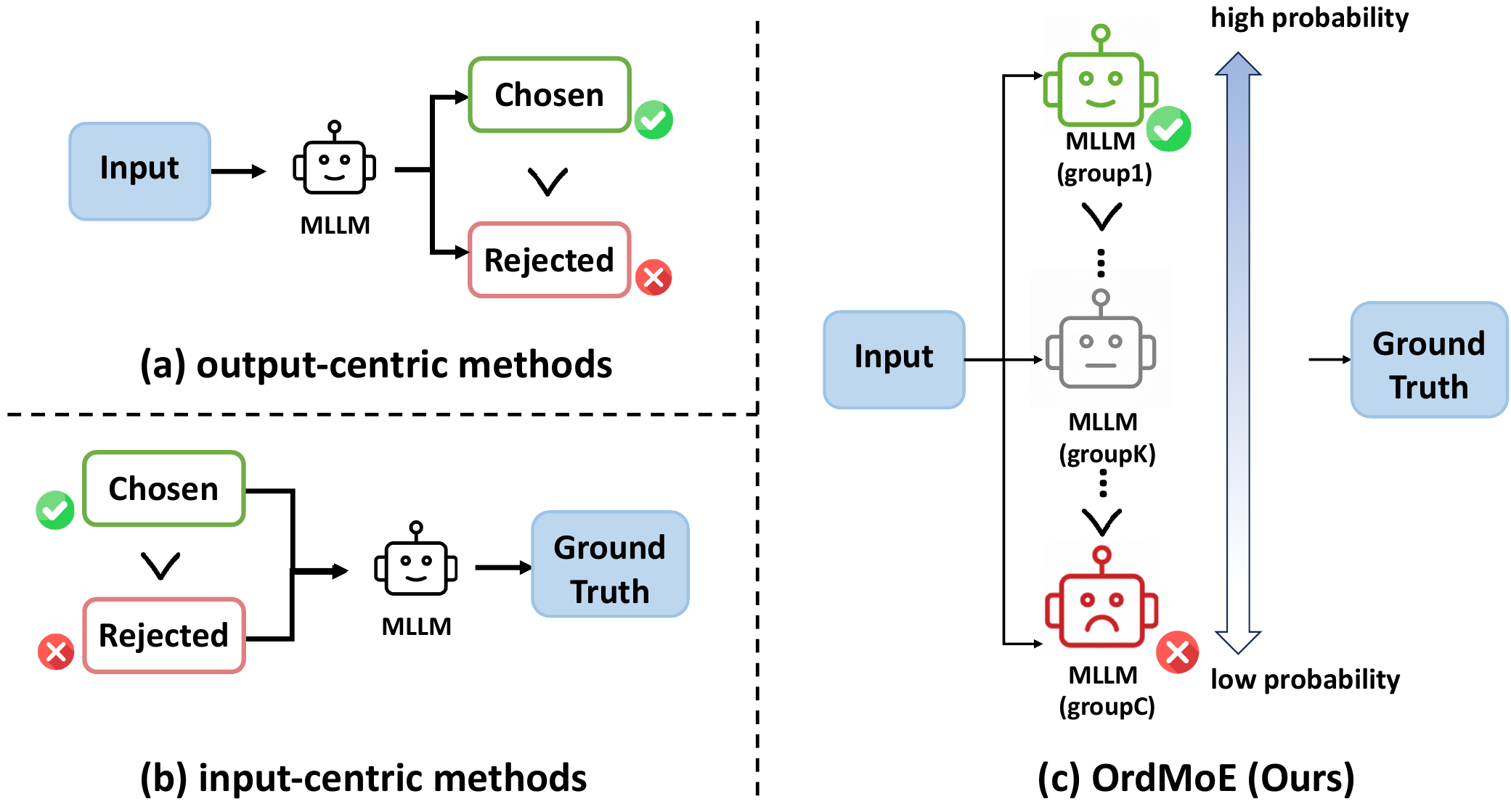}
    \caption{Comparison of preference learning paradigms: (a) output-centric (\textit{e.g.}, DPO) requires human-labeled preferences; (b) input-centric methods rely on prompt variations(\textit{e.g.}, mDPO); (c) our OrdMoE uses identical inputs and outputs, but exploits the MoE router’s intrinsic signal to construct a self-supervised expert ranking.}
    \label{fig:preferelearning_demo}
\end{figure}

Existing preference learning approaches for MLLMs rely on externally constructed preference signals, which fall into two categories. The first category, output-centric methods such as Direct Preference Optimization (DPO)~\cite{rafailov2023direct} and Reinforcement Learning from Human Feedback (RLHF)~\cite{ouyang2022training}, requires explicit human or teacher model judgments to label one response as preferred over another, making them costly and hard to scale (see Figure~\ref{fig:preferelearning_demo}(a)). The second category, input-centric methods, avoids human annotation by automatically generating preference pairs through input perturbations (\textit{e.g.}, randomly masking image regions or adding noise~\cite{wang2024mdpo}) and assuming that outputs from clean inputs are superior, as shown in Figure~\ref{fig:preferelearning_demo}(b). While fully automatic, input-centric approaches still depend on handcrafted transformation rules that may not reflect genuine multimodal quality and can introduce bias when the perturbation does not align with semantic meaning. In both cases, the preference signal is imposed from outside the model’s natural computation, rather than emerging from its internal dynamics.

In parallel, the Mixture-of-Experts (MoE) architecture has emerged as the de facto standard for scaling modern MLLMs~\cite{jiang2024mixtral, dai2024deepseekmoe, team2025kimi, ai2025ming}. By sparsely activating specialized subnetworks (experts) conditioned on input tokens, MoE enables substantial gains in both model capacity and computational efficiency. Critically, we observe that this \textbf{routing mechanism implicitly encodes a rich, fine-grained preference signal}. Specifically, the router’s expert selection logits reflect an internal assessment of which experts are most likely to generate high-quality outputs for a given token. This signal is zero-cost, naturally available during standard inference, and requires no external supervision or post-hoc annotation. To the best of our knowledge, this intrinsic signal from MoE routing has not been explored for alignment in MLLMs. We bridge this gap by proposing OrdMoE, as shown in Figure~\ref{fig:preferelearning_demo}(c), a novel preference alignment framework that directly harnesses the router’s expert selection logits as a self-supervised, token-level preference signal, eliminating the need for external human annotations or negative sampling. 

The core idea of OrdMoE is to transform the continuous routing scores into an internally consistent preference ordering. Specifically, we partition experts into ranked tiers based on their per-token activation logits and generate responses by activating each tier independently. This yields a sequence of outputs with progressively higher quality, effectively constructing a zero-cost preference hierarchy. The resulting ordered responses can then be optimized using standard preference learning objectives, enabling efficient, scalable, and fully self-supervised alignment for MoE-based MLLMs. We evaluate OrdMoE across multiple modalities and a diverse set of multimodal benchmarks, and our results consistently demonstrate its effectiveness.

Our contributions can be summarized as follows:

\begin{itemize}
    \item To the best of our knowledge, we present the first study to leverage the intrinsic preference signal inherently encoded within the Mixture-of-Experts (MoE) router’s expert selection scores for zero-cost preference alignment in MLLMs.
    \item We propose a novel training framework, OrdMoE, which constructs a multi-level, ordinal preference hierarchy among expert subsets based purely on their routing scores, thereby entirely bypassing the reliance on costly external human preference data.
    \item Extensive experiments across multiple multimodal benchmarks demonstrate that OrdMoE significantly enhances both alignment and overall performance of MoE-based multimodal large language models.
\end{itemize}

\section{Related Works}
\label{sec:Related}

\subsection{MoE-based MLLMs}

The Mixture-of-Experts (MoE)~\cite{jacobs1991adaptive} is a neural network architecture designed to scale model capacity through sparse activation and conditional computation. Mixtral 8x7B~\cite{jiang2024mixtral}, a prominent sparse MoE model based on the Mistral architecture, notably uses a router to select two out of eight experts per token at every layer. Subsequently, DeepSeek-MoE~\cite{dai2024deepseekmoe} further advanced the MoE framework by introducing the concepts of fine-grained experts and shared experts. The successful application of MoE in LLMs showcases the architecture's efficiency to achieve strong performance at a significantly lower computational expense.

Recently, some works  ~\cite{team2025kimi, lin2024moe, ai2025ming, li2025minimax} have integrated the MoE into MLLMs. For instance, Kimi-VL~\cite{team2025kimi} employs the Moonlight MoE language decoder, which has a total of 16B parameters but activates only 2.8B during inference. Qwen3-VL also adopts the MoE architecture for its language modeling component, based on the MoE design of Qwen3~\cite{yang2025qwen3}. Additionally, Ming-Omni~\cite{ai2025ming} introduces modality-specific routing heads that independently gate over a shared set of experts, allowing each modality to develop its own selection strategy without duplicating expert parameters. The fundamental rationale behind these designs is to leverage the sparsity of MoE to significantly scale the model's total parameter count, thereby achieving an optimal balance between inference efficiency and high performance across diverse tasks.

\subsection{Preference Learning}
Direct Preference Optimization (DPO)~\cite{rafailov2023direct} and Reinforcement Learning from Human Feedback (RLHF)~\cite{ouyang2022training} are the primary methods widely adopted in preference learning. However, recent works applying preference alignment to MLLM~\cite{sarkar2024data,yoon2025stop,peng2025mitigating,zhao2023beyond,zhou2024aligning,xiao2025detecting,wang2024mdpo} still rely on creating high-quality data. For instance,  mDPO~\cite{wang2024mdpo} generates hard negative (\textit{i.e.} rejected) samples by cropping 0–20\% of the original image, thereby producing non-preferred variants with reduced visual information. DPA~\cite{sarkar2024data} creates hallucinated samples using a "phrase-level modification" process to selectively alter object-related phrases in ground-truth correct responses. Our approach represents a radical departure by fundamentally leveraging the internal architecture of the MoE model to achieve preference alignment through an intrinsic architectural mechanism, entirely bypassing the need for external data generation or preference pair collection.

\section{Methodology}
\begin{figure*}[htbp] 
\centering
\includegraphics[width=1.0\textwidth, keepaspectratio]{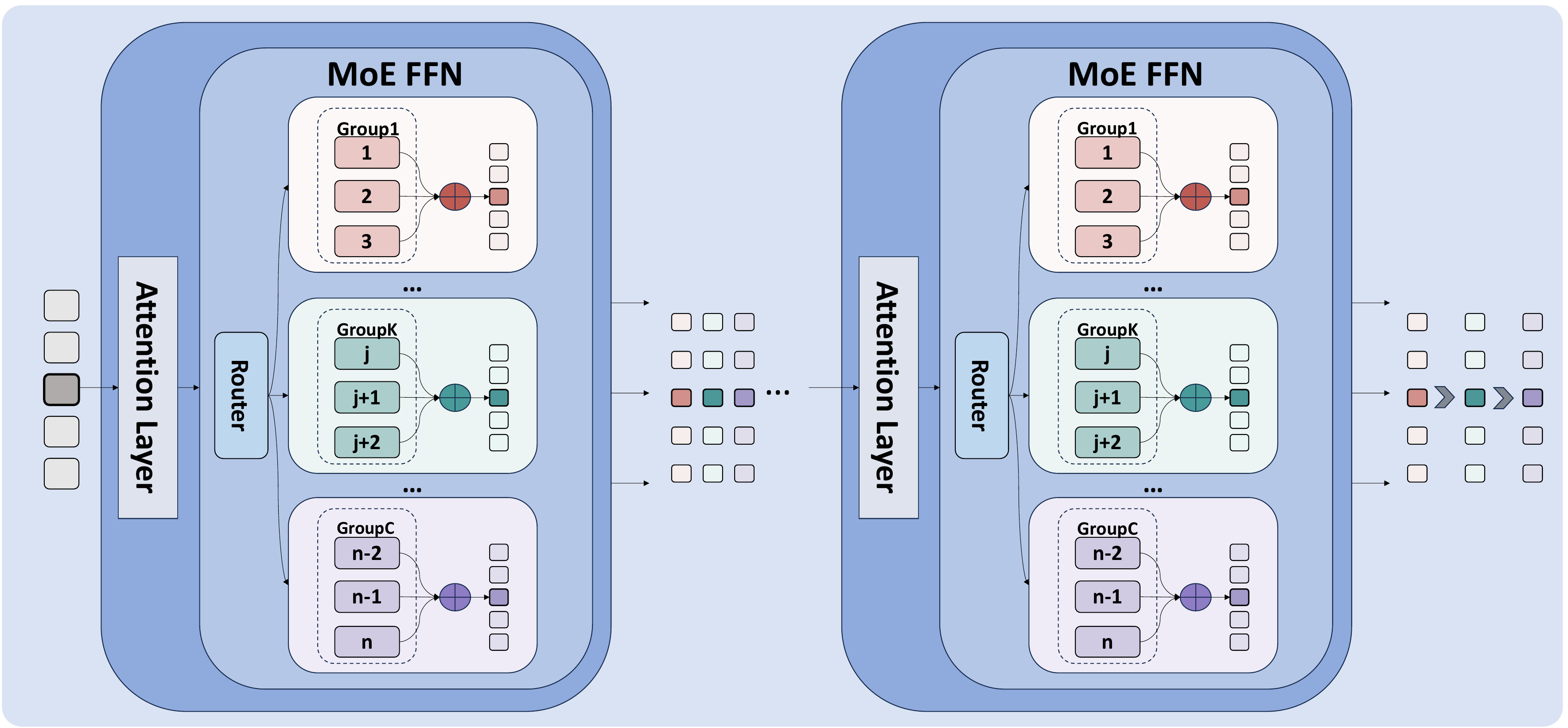}
\caption{Overall architecture of our OrdMoE training framework. The router computes routing probabilities and logically groups the selected experts into $\text{Group}_1 ...\text{Group}_K$... $\text{Group}_C$, corresponding to the highest, intermediate, and lower probabilities, respectively. The colored experts in each group denote this logical grouping: \textit{e.g.}, \textcolor[RGB]{234,200,198}{$\blacksquare$} (red-shaded) for $\text{Group}_1$, \textcolor[RGB]{171,206,205}{$\blacksquare$} (teal-shaded) for $\text{Group}_K$, and \textcolor[RGB]{208,203,227}{$\blacksquare$} (purple-shaded) for $\text{Group}_C$. Critically, the preference for the token are sequentially reduced from $\text{Group}_1$ to $\text{Group}_C$, directly reflecting the decreasing expert routing probabilities inherent to the OrdMoE design.}
\label{fig:method}
\end{figure*}
In this section, we introduce OrdMoE, a novel training framework for MoE-based multimodal large language models (MLLMs) that leverages the intrinsic properties of the MoE architecture to enable preference learning without relying on costly external human annotations.  As shown in Figure~\ref{fig:method}, the core of OrdMoE lies in  generating a set of $\mathbf{C}$ internally ranked response sequences. This is achieved by systematically constraining  expert selection in the MoE router at each layer $l$. Specifically, we select $\mathbf{C}$ ordered expert groups ($\mathbf{G}^{\pi_1}, \dots, \mathbf{G}^{\pi_{\mathbf{C}}}$) by partitioning the experts according to their routing scores, ranging from highest to lowest. By restricting each MoE layer to activate only the experts within a given group  $\mathbf{G}^{\pi_j}$, we produce $\mathbf{C}$ variants of the same input sequence, $\{\mathbf{y}^{\pi_1}, \mathbf{y}^{\pi_2}, \dots, \mathbf{y}^{\pi_{\mathbf{C}}}\}$. These responses exhibit an inherent ordinal preference relationship, with  $\mathbf{y}^{\pi_1}$ is the most preferred, and $\mathbf{y}^{\pi_{\mathbf{C}}}$ is the least preferred.

We now detail the components of OrdMoE: starting with preliminaries on MoE-based MLLMs (\cref{sec:preliminary}), followed by our intrinsic preference construction via routing-based expert grouping (\cref{sec:grouping}), constrained generation (\cref{sec:preference}), and the training objective(\cref{sec:loss}).

\subsection{Preliminary}
\label{sec:preliminary}

In MoE-based multimodal large language models (MLLMs), the input comprises tokens from multiple modalities.  Let $[\boldsymbol{V},\boldsymbol{A}, \boldsymbol{T}]$ denote the vision, audio, and text modalities, respectively. Their encoded representations are given by $\boldsymbol{V} \in \mathbb{R}^{L_V \times d}$, $\boldsymbol{A} \in \mathbb{R}^{L_A \times d}$, and $\boldsymbol{T} \in \mathbb{R}^{L_T \times d}$, obtained after processing through modality-specific encoders and projection layers. Here, $L_V, L_A, L_T$ are the modality-specific sequence lengths, and $d$ is the shared feature dimension. For any multimodal input, we concatenate the tokens from all modalities into a single sequence and feed it into the language model. Specifically, the input sequence at the first layer of the backbone is denoted as $\mathbf{X}$:
\begin{equation}
\mathbf{X} = [\mathbf{V} ; \mathbf{A} ; \mathbf{T}]= [\mathbf{x}_1, \mathbf{x}_2, \dots, \mathbf{x}_L] \in \mathbb{R}^{L \times d} 
\end{equation}
where $L = L_V + L_A + L_T$ is the total sequence length. Then, we pass this sequence of unified inputs through the LLM blocks. At each layer $l$, the representation of the $t$-th token, denoted $\mathbf{x}_t^l$, is routed to a dynamic subset of experts for processing. The router first computes a vector of expert logits $\boldsymbol{\rho}_{t}^l \in \mathbb{R}^n$, where n is the total number of experts:
\begin{equation}
	\boldsymbol{\rho}_{t}^l = \mathbf{W}_r \cdot \mathbf{x}_t^l + \mathbf{b}_r
\end{equation}
where $\mathbf{W}_r \in \mathbb{R}^{n \times d}$ and $\mathbf{b}_r \in \mathbb{R}^n$ are the learnable parameters of the router. Each entry $\rho_{i,t}^l$ denotes the logit corresponding to the $i$-th expert's logit for the $t$-th token.

The standard MoE selects the top-$K$ experts based on these logits. Let $\mathcal{K}(\mathbf{x}_t^l)$ denote the index set of the selected experts. The final routing weight $g_{i,t}^l$ for the $i$-th expert is obtained by applying  softmax normalization only over the selected subset:

\begin{equation}\label{equation3}
	g_{i,t}^l = \begin{cases} \frac{\exp\left( \rho_{i,t}^l \right)}{\sum_{j \in \mathcal{K}(\mathbf{x}_t^l)} \exp\left( \rho_{j,t}^l \right)} & \text{if } i \in \mathcal{K}(\mathbf{x}_t^l) \\ 0 & \text{otherwise} \end{cases}
\end{equation}
The layer's output $\mathbf{v}_t^l$ is computed as the sparsely weighted sum of the outputs from the selected experts:

\begin{equation}\label{equation4}
	\mathbf{v}_t^l = \sum_{i=1}^{n} g_{i,t}^l \cdot \mathbf{E}_i(\mathbf{x}_t^l)
\end{equation}

\subsection{Experts Grouping} 
\label{sec:grouping}
To construct a generalized intrinsic preference signal, we leverage the pre-computed routing scores $\rho_{i,t}^l$. Our approach introduces a single core hyperparameter: the number of preference groups, denoted $\mathbf{C}$. Each group consists of exactly $K$ experts (the same top-$K$ sparsity budget used in the original MoE model). From the full expert set  $\mathbf{E}$, we select $\mathbf{C}$ mutually exclusive expert subsets, $\mathbf{G}^{\pi_1}, \mathbf{G}^{\pi_2}, \dots, \mathbf{G}^{\pi_{\mathbf{C}}}$, each of size $K$. This selection induces a strict, ordered preference hierarchy over the $C$ groups, grounded in their routing scores. 

Specifically, for each token $t$ at layer $l$, we sort all experts $i \in \{1, \dots, n\}$ in descending order according to their normalized routing scores $\rho_{i,t}^l$. This sequential partitioning explicitly encodes a quality-ranked ordering of expert subsets, which serves as the basis for our self-supervised preference signal.

\subsection{Preference Definition} 
\label{sec:preference}

For a given input token embedding $\mathbf{x}$, we generate a sequence of $\mathbf{C}$ responses $\{\mathbf{y}^{\pi_1}, \mathbf{y}^{\pi_2}, \dots, \mathbf{y}^{\pi_{\mathbf{C}}}\}$, by constraining the MoE layer to activate experts only from the restricted group $\mathbf{G}^{\pi_j}$ along each activation path $\pi_j$, where $j \in \{1, 2, \dots, \mathbf{C}\}$. This controlled activation isolates the contribution of expert groups at different levels of routing probability, enabling a direct assessment of their relative output quality.

The restricted gating weight $\hat{g}_{i,t}^{l, \pi_j}$ for the $j$-th preference path is formally defined as the normalized routing score of expert $\mathbf{E}_i$ restricted to its assigned preference group $\mathbf{G}^{\pi_j}$ (Eq.~\ref{equation3}). Using these restricted weights, the output vector $\mathbf{v}_{t}^{l, \pi_j}$ is then computed by limiting the MoE operation to experts in $\mathbf{G}^{\pi_j}$ (Eq.~\ref{equation4}). This forward pass is applied consistently across all MoE layers and tokens, yielding a complete response $\mathbf{y}^{\pi_j}$ that reflects the collective capability of the $j$-th expert tier, with higher tiers expected to produce more accurate or coherent outputs due to their higher routing confidence.

This controllable generation process naturally induces a ranked sequence of responses, ordered by the experts’ initial routing probabilities:
\begin{equation}
\mathbf{y}^{\pi_1} \succ \mathbf{y}^{\pi_2} \succ \dots \succ \mathbf{y}^{\pi_{\mathbf{C}}}
\end{equation}

The response $\mathbf{y}^{\pi_1}$,  generated by the experts with the highest routing probabilities, is treated as the most preferred output, whereas $\mathbf{y}^{\pi_{\mathbf{C}}}$, produced by those with the lowest routing probabilities, is regarded as the least preferred. This induces a strict, multi-level preference hierarchy, which we map to a fixed intrinsic reward signal $r^{\pi_j}$ satisfying:
\begin{equation}
\mathbf{{r}}^{\pi_1} \succ \mathbf{r}^{\pi_2} \succ \dots \succ \mathbf{r}^{\pi_{\mathbf{C}}}
\end{equation}

This structured reward enables the model to learn from fine-grained distinctions across the $\mathbf{C}$ tiers of expert quality.
\subsection{Training Objective}
\label{sec:loss}

The overall training objective consists of three components: (1) a ranking loss applied to expert groups, (2) the standard next-token prediction loss, and (3) a load-balancing loss to encourage equitable utilization of experts.

\paragraph{Expert Rank Loss}
\label{sec:rank_loss}
The objective of this optimization is to induce a discriminative ranking among experts, such that their output distributions align with the ground-truth distribution in accordance with their assigned preference scores. This strategy enforces a clear separation in the output behaviors of the $\mathbf{C}$ distinct expert groups, reflecting their intrinsic ranking hierarchy.

The average token log-probability for a response variant $\mathbf{y}^{\pi_j}$ is defined as: 
\begin{equation}
\bar{\mathcal{L}}^{\pi_j} = \frac{1}{|\mathbf{y}^{\pi_j}|} \sum_{t=1}^{|\mathbf{y}^{\pi_j}|} \log \pi (\mathbf{y}_t^{\pi_j} | \mathbf{y}_{<t}^{\pi_j})
\end{equation}

We begin by assigning a fixed, monotonically decreasing intrinsic reward $r^{\pi_j}$ to each of the $\mathbf{C}$ response paths, such that  $r^{\pi_1} > r^{\pi_2} > \dots > r^{\pi_{\mathbf{C}}}$. To stabilize training, we compute the advantage function $\mathcal{A}^{\pi_j}$ by applying batch-wise z-score normalization to the set of intrinsic rewards $r^{\pi_1} > r^{\pi_2} > \dots > r^{\pi_{\mathbf{C}}}$: 
\begin{equation}
\mathcal{A}^{\pi_j} = \frac{r^{\pi_j} - \mu_r}{\sigma_r}
\end{equation}
where $\mu_r$ and $\sigma_r$ denote the mean and standard deviation of the $\mathbf{C}$ assigned rewards within the current training batch.

The final expert rank Loss ($\mathcal{L}_{\text{ERL}}$) is formulated as a generalized policy gradient objective. Specifically, we minimize the negative expected weighted sum of average token log-probabilities across all $\mathbf{C}$ generated sequence variants:
\begin{equation}
\mathcal{L}_{\text{ERL}} = - \mathbb{E}_{\mathbf{x}, \{\mathbf{y}^{\pi_j}\} \sim \pi} \left[ \sum_{j=1}^{\mathbf{C}} \mathcal{A}^{\pi_j} \cdot \bar{\mathcal{L}}^{\pi_j} \right]
\end{equation}

By minimizing $\mathcal{L}_{\text{ERL}}$, the model is encouraged to generate outputs that align with the high-confidence decisions of its routing mechanism, thereby effectively exploiting the intrinsic preference hierarchy among expert groups. 

\paragraph{Next Token Prediction}
The primary objective remains the standard next-token prediction loss ($\mathcal{L}_{\text{NTP}}$), which preserves the model’s core autoregressive generative capability. Specifically, $\mathcal{L}_{\text{NTP}}$ is computed as the cross-entropy between the predicted next-token distribution and the ground-truth tokens, using only the output of the highest-ranked expert group ($\mathbf{G}^{\pi_1}$). This design ensures that the generative backbone is trained exclusively on the most confident routing decision, aligning with conventional MoE practice where only the selected experts contribute to the final output and loss computation.

\begin{equation}\label{eq:loss_ntp}
    \mathcal{L}_{\text{NTP}} = \mathcal{L}_{\text{CE}}(\mathbf{y}^{\pi_1}, \mathbf{y}_{\text{gt}})
\end{equation}

Here, $\mathbf{y}^{\pi_1}$ denotes the sequence generated by the top-ranked expert group $\mathbf{G}^{\pi_1}$, and $\mathbf{y}_{\text{gt}}$ is the corresponding ground-truth target sequence. 

\paragraph{Balancing Loss}

To promote efficient expert utilization and prevent expert collapse, we incorporate the standard Mixture-of-Experts (MoE) balancing loss ($\mathcal{L}_{\text{balance}}$). Given our architecture generates multiple response paths $\{\mathbf{y}^{\pi_1}, \mathbf{y}^{\pi_2}, \dots, \mathbf{y}^{\pi_{\mathbf{C}}}\}$, this loss is applied to the primary output path $\mathbf{y}^{\pi_1}$ to enhance its stability and reliability. Specifically, the loss is computed to uniformly distribute the workload across all experts $\mathbf{E}$, as:

\begin{equation}\label{eq:loss_balance}
\mathcal{L}_{\text{balance}} = \sum_{j \in \mathbf{E}} p_{j} \cdot f_{j}
\end{equation}

where $p_j$ denotes the mean routing probability assigned to expert $\mathbf{E}_j$, computed as $p_j = \frac{1}{L} \sum_{i=1}^{L} P(e=j | x_i)$, and $f_j$ represents the fraction of total expert capacity utilized by $\mathbf{E}_j$, given by $f_j = \frac{c_j}{L}$. Here $L$ is the total number of tokens in the batch, and $c_j$ is the number of tokens routed to expert $\mathbf{E}_j$, where only tokens assigned based on $\pi_1$ (the first routing path) are counted for $c_j$.

\paragraph{Total Training Objective}
The final training objective integrates three components: 

\begin{equation}\label{eq:total_loss_final}
    \mathcal{L}_{\text{total}} = \mathcal{L}_{\text{NTP}} + \lambda_{\text{ERL}} \mathcal{L}_{\text{ERL}} + \lambda_{\text{balance}} \mathcal{L}_{\text{balance}}
\end{equation}

where, $\lambda_{\text{ERL}}$ and $\lambda_{\text{balance}}$ are hyperparameter coefficients that control the influence of the expert rank loss and balancing loss, respectively. We set both to 1.0 in all experiments.

\section{Experiments}
\subsection{Experiments Setup}
\subsubsection{Model Configuration}
In this work, we adopt Ming-Lite-Omni~\cite{ai2025ming} (a powerful, open-source, pre-trained multimodal large language model) as our primary baseline and the foundation for all main experiments. Its state-of-the-art performance across a wide range of multimodal benchmarks provides a strong and reliable starting point for our study. For ablation purposes, we also conduct supplementary experiments using Ling-mini-2.0~\cite{ling-mini}, a smaller-scale alternative, to assess the robustness and generalizability of our approach across different model scales. 

In our OrdMoE framework, the number of expert groups is set to $\mathbf{C}=3$. To construct these groups, we first sort all experts in the base model according to their average routing scores, then partition the sorted list into contiguous blocks. For Ming-Lite-Omni, which has 64 total experts with top-$K$=6, we form three groups of 6 experts each by selecting the top block (ranks 1–6), a middle block (ranks 25–30, \textit{i.e.}, the 5th blocks of size 6), and the bottom block (ranks 59–64). For Ling-mini-2.0, which contains 256 experts with top-$K$=8 routing, we similarly divide the sorted expert list into 32 blocks of 8 experts and select the first (ranks 1–8), the 16th (ranks 121–128), and the last (ranks 249–256) blocks. This uniform sampling strategy ensures that the selected groups span the full spectrum of expert specializations (from high-confidence to low-confidence) enabling meaningful ranking supervision in $\mathcal{L}_{\text{ERL}}$.

\subsubsection{Training Details}
Our model is trained under two distinct data regimes, both sharing the same architecture and optimization protocol to enable a fair comparison of how data composition affects performance and generalization. The first regime uses a vision–language-only corpus, consisting of:
\begin{itemize}
\item Pretraining: 6 million image–text pairs and 3 million text-only language understanding corpus;
\item Supervised Fine-Tuning: 9 million multimodal question-answering pairs and 3 million text-only NLP instructions.
\end{itemize}

The second regime, more comprehensive regime employs a full-spectrum multimodal corpus, incorporating diverse data from text, image, audio, and video modalities (details in Appendix). Both stages (pretraining and SFT) are correspondingly extended to this heterogeneous setting, ensuring consistent training dynamics across modalities.

For the experiments based on Ming-Lite-Omni,  we initialize from its multimodally pretrained weights, where image-text alignment has already been established through prior pretraining. We then apply continued pretraining (under either the vision-language or full-modality corpus) followed by supervised fine-tuning (SFT) within the OrdMoE framework. To further evaluate the adaptability of OrdMoE to different language model backbones, we also instantiate it on Ling-mini-2.0. Starting from its original pretrained LLM weights, we perform full multimodal adaptation via pretraining and SFT, applying $\mathcal{L}_{\text{ERL}}$ throughout both stages. 

\subsubsection{Evaluation Datasets}
Our model is evaluated across a diverse set of audio, image, and video understanding tasks to comprehensively assess its multimodal perception and reasoning capabilities. For image-text understanding, we select a suite of challenging multimodal and vision-centric benchmarks:
AI2D~\cite{kembhavi2016diagram} (diagram understanding),
MMMU~\cite{yue2024mmmu} (massive multi-discipline multimodal understanding),
MMStar~\cite{chen2024we} (complex multimodal reasoning),
OCRBench~\cite{liu2024ocrbench} (OCR robustness),
MM-Vet~\cite{yu2023mm}(comprehensive multimodal capability), and
Mathvista~\cite{lu2023mathvista} (mathematical reasoning in visual contexts). For video understanding, we include  LongVideoBench~\cite{wu2024longvideobench}, DiDeMo~\cite{anne2017localizing}, AVQA~\cite{yang2022avqa}, MVBench~\cite{li2024mvbench}, and Video-MME~\cite{fu2025video}, using a uniform sampling strategy that extract 128 frames per video. For audio performance, we evaluate the model on standard speech recognition benchmarks, including public Chinese datasets such as Aishell1~\cite{bu2017aishell} and English datasets such as LibriSpeech~\cite{zeyer2021librispeech}. This selection spans a broad spectrum of task difficulties, modalities, and domains, enabling a holistic evaluation of the model’s fundamental perceptual abilities and advanced reasoning skills.

\subsection{Results on Vision-language models}
\label{exp:vlm}
We begin by evaluating OrdMoE on Ming-Lite-Omni, initializing from its multimodally pretrained weights and performing continued pretraining (using vision–language data) followed by supervised fine-tuning.  As shown in Table~\ref{tab:vlm}, both the baseline and OrdMoE use identical data, initialization, and training pipelines—the sole distinction being the inclusion of the expert rank Loss ($\mathcal{L}_{\text{ERL}}$) in OrdMoE. Despite this minimal modification to the training objective, OrdMoE yields consistent and substantial performance gains over the baseline across all benchmarks, underscoring the value of harnessing intrinsic routing signals for self-supervised preference learning. 
\begin{table}[H]
    \centering
    \caption{OrdMoE vs. SFT baseline on six image understanding benchmarks. Models differ only in the use of $\mathcal{L}_{\text{ERL}}$.}
    \label{tab:vlm}
    \scriptsize
    \renewcommand{\arraystretch}{1.1} 
    \setlength{\tabcolsep}{1.4pt} 
    \begin{tabular}{@{}l *{7}{S[table-format=2.3, retain-explicit-plus]} @{}} 
        \toprule
        & \multicolumn{7}{c}{Image Understanding} \\
        \cmidrule(lr){2-8} 
        Model 
        & {\makecell{AI2D}} 
        & {\makecell{MMMU}}
        & {\makecell{MMStar}} 
        & {\makecell{OCRBench}} 
        & {\makecell{MMVet}} 
        & {\makecell{Mathvista}} 
        & {\makecell{Avg $\uparrow$}} \\ 
        \midrule
        Baseline      & 69.40 & 40.44 & 45.03 & 76.20 & 55.27 & 51.60 & 56.32 \\ 
        OrdMoE        & 77.10 & 48.11 & 54.98 & 77.80 & 59.04 & 59.50 & 62.75 \\ 
     \rowcolor{gray!20}
     Improve($\Delta$) & \textcolor{ForestGreen}{+7.70} & \textcolor{ForestGreen}{+7.67} & \textcolor{ForestGreen}{+9.95} & \textcolor{ForestGreen}{+1.60} & \textcolor{ForestGreen}{+3.77} & \textcolor{ForestGreen}{+7.90} & \textcolor{ForestGreen}{+6.43} \\
     \bottomrule
    \end{tabular}
\end{table}
\subsection{Results on Omni-models}
In contrast to the vision–language-only setting in Section~\ref{exp:vlm}, this section evaluates OrdMoE under a full-modality training regime that incorporates image, audio, and video data, enabling assessment in a more comprehensive multimodal scenario. As shown in Table~\ref{tab:omni}, OrdMoE consistently improves over the baseline across all modalities: it reduces word error rate (WER) on audio benchmarks (lower is better), and increases accuracy on both image and video understanding tasks (higher is better). These results demonstrate that the proposed expert ranking mechanism generalizes robustly beyond vision–language tasks and remains effective when trained on heterogeneous, full-spectrum multimodal data.

\subsection{Ablation Studies}
We conduct comprehensive ablation studies to thoroughly investigate the impact of key design choices in OrdMoE. Specifically, we examine three critical aspects: (1) the number of preference groups ($C$), (2) the expert grouping strategy—namely, how experts are assigned to ranked tiers based on their routing behavior, (3) the magnitude of the group-level preference reward ($r$), and (4) the layer scope over which the group-level preference is utilized. These experiments collectively shed light on the sensitivity of our method to architectural and hyperparameter configurations, and validate the effectiveness of our proposed ordinal experts grouping scheme.
\begin{table}[H]
    \centering
    \caption{Generalization of OrdMoE to omni-modality training: consistent gains over the baseline across audio, image, and video understanding tasks. (Note: VMME denotes VideoMME, and VMME(*) indicates evaluation with subtitles.)}
    \label{tab:omni}
    \scriptsize
    \renewcommand{\arraystretch}{1.1} 
    \setlength{\tabcolsep}{0.8pt} 
    \begin{tabular}{@{}l *{7}{S[table-format=2.3]} @{}} 
        \toprule
        & \multicolumn{7}{c}{{Audio Understanding}} \\
        \cmidrule(lr){2-8} 
        {Model} 
        & {\makecell{Aishell1}}
        & {\makecell{Aishell2\\(Android)}}
        & {\makecell{Aishell2\\(iOS)}}
        & {\makecell{Libri\\(Clean)}}
        & {\makecell{Libri\\(Other)}}
        & {\makecell{MultiLibri}}
        & {\makecell{Avg $\downarrow$}} \\
        \midrule

        {Baseline} & 1.78 & 2.97 & 2.99 & 1.61 & 3.63 & 5.17 & 3.03 \\
        {OrdMoE} & 1.51 & 2.66 & 2.64 & 1.43 & 3.12 & 5.28 & 2.77 \\
        \rowcolor{gray!20} Improve($\Delta$) & \textcolor{ForestGreen}{+0.27} & \textcolor{ForestGreen}{+0.31} & \textcolor{ForestGreen}{+0.35} & \textcolor{ForestGreen}{+0.18} & \textcolor{ForestGreen}{+0.51} & -0.11 & \textcolor{ForestGreen}{+0.26} \\
        \bottomrule
        \toprule 
        
        & \multicolumn{7}{c}{{Image Understanding}} \\
        \cmidrule(lr){2-8}
        & {\makecell{AI2D}} 
        & {\makecell{MMMU}}
        & {\makecell{MMStar}} 
        & {\makecell{OCRBench}} 
        & {\makecell{MMVet}} 
        & {\makecell{Mathvista}} 
        & {\makecell{Avg $\uparrow$}} \\ 
        \midrule

        {Baseline} & 72.51 & 45.67 & 55.00 & 79.30 & 60.00 & 56.60 & 61.51 \\ 
        {OrdMoE} & 74.90 & 46.89 & 56.58 & 80.10 & 57.29 & 58.50 & 62.38 \\ 
        \rowcolor{gray!20} Improve($\Delta$) & \textcolor{ForestGreen}{+2.39} & \textcolor{ForestGreen}{+1.22} & \textcolor{ForestGreen}{+1.58} & \textcolor{ForestGreen}{+0.80} & -2.71 & \textcolor{ForestGreen}{+1.90} & \textcolor{ForestGreen}{+0.88} \\
        \bottomrule
        \toprule

        & \multicolumn{7}{c}{{Video Understanding}} \\
        \cmidrule(lr){2-8}
        & {\makecell{VMME}} 
        & {\makecell{$\text{VMME}^{*}$}}
        & {\makecell{MVBench}} 
        & {\makecell{LongVideo}} 
        & {\makecell{DiDeMo}} 
        & {\makecell{AVQA}} 
        & {\makecell{Avg $\uparrow$}} \\  

        \midrule
        {Baseline} & 54.96 & 61.81 & 53.53 & 50.11 & 41.20 & 82.65 & 57.38 \\ 
        {OrdMoE} & 56.85 & 63.22 & 57.10 & 51.76 & 43.87 & 84.67 & 59.58 \\ 
        \rowcolor{gray!20} Improve($\Delta$) & \textcolor{ForestGreen}{+1.89} & \textcolor{ForestGreen}{+1.41} & \textcolor{ForestGreen}{+3.57} & \textcolor{ForestGreen}{+1.65} & \textcolor{ForestGreen}{+2.67} & \textcolor{ForestGreen}{+2.02} & \textcolor{ForestGreen}{+2.20} \\
        \bottomrule
    \end{tabular}
\end{table}
\subsubsection{The impact of preference group number}

To determine the optimal grouping complexity, we vary the number of preference groups $C$ while keeping all other hyperparameters constant. Specifically, for Ming-Lite-Omni, which employs $|\mathbf{E}|=64$ experts with top-$K=6$ routing, we first sort all experts by their routing score and then select $C$ non-overlapping groups of 6 experts each by uniformly sampling blocks from the ranked list, always including the top (ranks 1–6) and bottom (ranks 59–64) blocks when $C\geq2$. Concretely:

\begin{itemize}
    \item For $C=1$, only a single group is used (effectively disabling ordinal preference learning). 
    \item For $C=2$, we use the top block (ranks 1–6) and the bottom block (ranks 59–64); 
    \item For $C=3$, we additionally include a middle block (ranks 25–30, \textit{i.e.}, the 5th block);
    \item For $C=4$, we further sample two intermediate blocks (ranks 19–24 and 37–42), resulting in four evenly spaced tiers.
\end{itemize}

This design allows us to systematically study how the granularity of the intrinsic preference hierarchy affects alignment performance. As shown in the average column of Table~\ref{tab:ablation_group_c}, which averages performance across six multimodal benchmarks, the model achieves 56.32\% with $C=1$, corresponding to the SFT baseline (\textit{i.e.}, training without ordinal preference supervision). Introducing a two-tier hierarchy ($C=2$) yields only marginal gain (56.42\%). In contrast, $C=3$ leads to a substantial improvement to 62.75\%, while $C=4$ slightly drops to 61.78\%. The slight degradation at $C=4$ further suggests that excessive granularity may dilute the preference signal or create competition between alignment and generative objectives. Thus, $C=3$ emerges as the sweet spot for intrinsic preference learning in MoE-based MLLMs.

\begin{table}[H]
    \centering
    \scriptsize 
    \renewcommand{\arraystretch}{1.1} 
    \setlength{\tabcolsep}{2.0pt} 
    \caption{Effect of preference group $C$ on multimodal performance.}
    \label{tab:ablation_group_c}
    \begin{tabular}{l|ccccccc}
        \toprule
        \textbf{C} 
        & AI2D & MMMU & MMStar & OCRBench & MMVet & Mathvista & Avg $\uparrow$ \\ \midrule
        1(Baseline) & 69.40 & 40.44 & 45.03 & 76.20 & 55.27 & 51.60 & 56.32 \\ 
        2 & 70.89 & 39.56 & 47.72 & 74.40 & 52.89 & 53.07 & 56.42 \\ 
        \rowcolor{gray!20} 3(OrdMoE) & 77.10 & 48.11 & 54.98 & 77.80 & 59.04 & 59.50 & 62.76 \\ 
        4 & 75.87 & 47.67 & 54.16 & 78.10 & 57.98 & 56.93 & 61.78 \\ 
        \bottomrule
    \end{tabular}
\end{table}

\subsubsection{The impact of grouping strategy}

The second key design choice in OrdMoE is how experts are assigned to preference groups. Our default strategy uniformly samples non-overlapping blocks of size $K=6$ from the routing-score-ranked expert list (\textit{e.g.}, top, middle, and bottom blocks for $C=3$), ensuring a broad coverage of quality levels. To validate the necessity of this structured diversity, we compare against two alternative schemes: 

\begin{itemize}  
    \item High-only: all groups are drawn from the highest-ranking experts (\textit{e.g.}, ranks 1–6, 7–12, 13–18), resulting in minimal quality separation between tiers;
    \item Random grouping: experts are shuffled uniformly at random before partitioning into contiguous blocks of size 6, thereby discarding the routing-score ordering entirely. 
\end{itemize}

As shown in Table~\ref{tab:ablation_group_strategy}, both the high-only and random grouping strategies underperform significantly compared to uniform sampling. The high-only variant achieves a moderate drop (61.28\% vs. 62.76\%), indicating that insufficient quality separation between tiers limits the effectiveness of ordinal supervision. In stark contrast, random grouping, despite using the same number of groups, collapses to near-baseline performance (56.67\%), demonstrating that the ordinal structure derived from routing behavior, not merely the presence of multiple groups, is essential for meaningful preference learning.

\begin{table}[H]
    \centering
    \scriptsize 
    \renewcommand{\arraystretch}{1.1} 
    \setlength{\tabcolsep}{2.0pt} 
    \caption{Ablation on expert grouping strategy. Uniform sampling spans the full routing-quality spectrum; high-only lacks inter-tier diversity; random discards ranking. OrdMoE employs the uniform sampling strategy. }
    \label{tab:ablation_group_strategy}
    \begin{tabular}{l|ccccccc}
        \toprule
        \textbf{Strategy} 
        & AI2D & MMMU & MMStar & OCRBench & MMVet & Mathvista & Avg $\uparrow$ \\ \midrule
        Baseline & 69.40 & 40.44 & 45.03 & 76.20 & 55.27 & 51.60 & 56.32 \\ 
        \rowcolor{gray!20} Uniform & 77.10 & 48.11 & 54.98 & 77.80 & 59.04 & 59.50 & 62.76 \\
        High-only & 75.71 & 46.33 & 53.66 & 76.70 & 57.61 & 57.67 & 61.28 \\
        Random & 69.17 & 43.00 & 49.64 & 73.80 & 54.31 & 50.07 & 56.67 \\
        \bottomrule
    \end{tabular}
\end{table}

A complementary design question is whether the size of each preference group should align with the number of activated experts per token ($K=6$). In our default (\textit{even}) setting, each of the $C=3$ groups contains exactly $K=6$ experts, matching the router’s activation capacity so that every tier represents a complete routing configuration. We contrast this with an \textit{uneven} variant that uses heterogeneous group sizes: the top tier retains 6 experts (ranks 1–6), while the second and third tiers each contain only 3 experts (\textit{e.g.}, ranks 25–27 and 54–56). This investigates whether high-quality ordinal supervision requires each tier to constitute a complete top-$K$ expert set (as used during MoE training), rather than sparse or disjoint subsets of experts that lack coherent collaborative dynamics. 

Results in Table~\ref{tab:ablation_group_uneven} show that the even grouping (6 experts per tier) yields a modest but consistent improvement over the uneven variant (Tier 1: 1-6, Tier 2: 25-27, Tier 3: 59–61). We attribute this to the fact that the underlying MoE language model is trained with top-$K$ = 6 routing, making groups of size $K$ natural units of expert co-activation. Aligning preference tiers with this native granularity thus provides more coherent supervision during ordinal learning. Consequently, OrdMoE adopts the even grouping strategy by default.

\begin{table}[H]
    \centering
    \scriptsize 
    \renewcommand{\arraystretch}{1.1} 
    \setlength{\tabcolsep}{2.0pt} 
    \caption{Effect of uniform v.s. non-uniform group sizes in ordinal tiers ($C=3$). \textit{Even}: all tiers use 6 experts (aligned with top-$K=6$). \textit{Uneven}: Tier 1 uses 6 experts; Tiers 2–3 use 3 experts each.}
    \label{tab:ablation_group_uneven}
    \begin{tabular}{l|ccccccc}
        \toprule
        \textbf{Strategy} 
        & AI2D & MMMU & MMStar & OCRBench & MMVet & Mathvista & Avg $\uparrow$ \\ \midrule
        \rowcolor{gray!20} Even(OrdMoE) & 67.75 & 40.11 & 49.58 & 74.10 & 58.12 & 48.90 &  56.43\\ 
        Uneven & 65.32 & 38.22 & 48.79 & 75.10 & 52.61 & 48.23 &  54.71\\  
        \bottomrule
    \end{tabular}
\end{table}
 
\subsubsection{The impact of group preference reward}

The results in Table~\ref{tab:ablation_reward} reveal a clear sensitivity to the scale of the group-level reward. While all three reward schemes encode the same ordinal ranking (high > medium > low), they differ in the magnitude of incentive separation between tiers. The default setting $r=[1.0, 0.5, 0]$ achieves the highest average score (62.76\%), outperforming both stronger ($[2,1,0]$, 60.89\%) and weaker ($[0.5,0.25,0]$, 60.36\%) variants. This indicates that moderate reward scaling strikes the optimal balance: it provides sufficient pressure to encourage clear quality-based differentiation among expert groups, without overwhelming the primary language modeling objective or causing optimization instability. In contrast, excessively large rewards ($[2,1,0]$) disrupt the balance between preference learning and language modeling, likely by over-suppressing mid-tier experts, whereas overly small rewards ($[0.5,0.25,0]$) provide insufficient gradient signal to enforce meaningful quality ordering. These findings confirm that the relative spacing of rewards, not just their ordering, plays a critical role in self-supervised preference learning.

\begin{table}[H]
    \centering
    \scriptsize 
    \renewcommand{\arraystretch}{1.1} 
    \setlength{\tabcolsep}{1.0pt} 
    \caption{Ablation on group-level preference reward magnitude $r$ with $C=3$.}
    \label{tab:ablation_reward}
    \begin{tabular}{l|ccccccc}  
        \toprule
        \textbf{Reward} 
        & AI2D & MMMU & MMStar & OCRBench & MMVet & Mathvista & Avg $\uparrow$ \\ \midrule
         \rowcolor{gray!20}{[1,0.5,0] (OrdMoE)}   & 77.10 & 48.11 & 54.98 & 77.80 & 59.04 & 59.50 & 62.76 \\ 
        {[2,1,0]}     & 75.03 & 47.22 & 53.12 & 77.90 & 56.51 & 55.53 & 60.89 \\
        {[0.5,0.25,0]}& 73.51 & 45.22 & 51.03 & 75.20 & 61.06 & 56.17 & 60.36 \\ \bottomrule
    \end{tabular}
\end{table}

\subsubsection{The impact of layer scope}

To investigate how the placement of group-level preference signals affects model performance, we conduct a series of ablation studies within our OrdMoE framework. Our model comprises 28 layers in total. We evaluate four configurations that differ in the layer scope over which the OrdMoE mechanism is applied. 

\begin{itemize}
    \item Full layers (OrdMoE): all layers adopt the OrdMoE mechanism, constructing three group-level preference signals from the \(C=3\) uniformed ranked expert groups; 
    \item Shallow layers: the OrdMoE mechanism is active only in the first four layers (1–4);
    \item Deep layers: the OrdMoE mechanism is active only in the last four layers (25–28);
    \item Even layers: the OrdMoE mechanism is active only in four sparsely sampled layers: 7, 14, 21, and 28.
\end{itemize}

In all ablated variants, layers outside the specified scope employ the conventional MoE expert selection, choosing the top-$K$ experts with the highest routing probabilities.

\begin{table}[H]
    \centering
    \scriptsize 
    \renewcommand{\arraystretch}{1.1} 
    \setlength{\tabcolsep}{1.0pt}  
    \caption{Ablation study on the layer scope of the OrdMoE mechanism \(C=3\). We compare applying group-level ordinal routing in all layers (Full) versus only in shallow, deep, or sparsely sampled layers.}
    \label{tab:ablation_layer}
    \begin{tabular}{l|ccccccc}  
        \toprule
        \textbf{Layer Scope} 
        & AI2D & MMMU & MMStar & OCRBench & MMVet & Mathvista & Avg $\uparrow$ \\ \midrule
         \rowcolor{gray!20}{Full layers (OrdMoE)}   & 77.10 & 48.11 & 54.98 & 77.80 & 59.04 & 59.50 & 62.76 \\ 
        {Shallow layers}     & 75.49 & 47.11 & 56.07 & 77.00 & 58.53 & 56.93 & 61.86 \\
        {Deep layers}& 76.68 & 44.00 & 53.88 & 77.00 & 56.60 & 55.17 &  60.56\\ 
        {Even layers}& 75.70 & 47.40 & 53.86 & 79.30 & 59.54 & 56.00 & 61.97 \\\bottomrule
    \end{tabular}
\end{table}

As shown in Table~\ref{tab:ablation_layer}, all partial-layer variants (shallow layers, deep layers, even layers) underperform the full-layer OrdMoE, highlighting the value of leveraging group-level preferences across the entire network. The shallow variants show a 0.9\% performance drop and the deep layer variants exhibit a more substantial 2.2\% performance drop , indicating that confining the ordinal preference mechanism to a limited layer range undermines the consistency of expert selection guidance. In contrast, the even-layer variant achieves a relatively milder degradation (61.97\% vs. 62.76\%), suggesting that sparse but strategically distributed layer involvement can partially preserve the benefits of ordinal supervision. Collectively, these results demonstrate that the full-layer scope, rather than partial or sparse layer application, is critical for maximizing the effectiveness of the group-level preference learning in OrdMoE.

\subsection{Results on different language model}

To evaluate the adaptability of OrdMoE to diverse language model foundations, we construct a multimodal variant using \textit{Ling-mini-2.0} (a sparse, open-source language model distinct from our primary base model) as the interface. Starting from its original pretrained LLM weights (without any prior multimodal alignment), we perform multimodal adaptation: first aligning the language model with a vision encoder via pretraining on image–text data, followed by supervised fine-tuning (SFT) on multimodal instruction tuning data. OrdMoE is applied throughout both stages. 

As shown in Table~\ref{tab:lingmini}, this approach yields consistent gains over the SFT-only baseline (+8.74\% average), despite using identical initialization, architecture, and training data. This confirms that $\mathcal{L}_{\text{ERL}}$ provides a plug-and-play enhancement that generalizes across different language model backbones, without requiring model-specific tuning or pre-adapted checkpoints.

\begin{table}[H]
    \centering
    \caption{OrdMoE vs. SFT baseline on six image understanding benchmarks, both trained from the same \textit{Ling-mini-2.0} LLM checkpoint through full multimodal alignment (pretraining + SFT). The only difference is the inclusion of $\mathcal{L}_{\text{ERL}}$.}
    \label{tab:lingmini}
    \scriptsize
    \renewcommand{\arraystretch}{1.1} 
    \setlength{\tabcolsep}{1.4pt} 
    \begin{tabular}{@{}l *{7}{S[table-format=2.3, retain-explicit-plus]} @{}} 
        \toprule
        & \multicolumn{7}{c}{Image Understanding} \\
        \cmidrule(lr){2-8} 
        Model 
        & {\makecell{AI2D}} 
        & {\makecell{MMMU}}
        & {\makecell{MMStar}} 
        & {\makecell{OCRBench}} 
        & {\makecell{MMVet}} 
        & {\makecell{Mathvista}} 
        & {\makecell{Avg $\uparrow$}} \\ 
        \midrule
        Baseline      & 55.57 & 33.56 & 30.63 & 60.90 & 45.50 & 39.13 & 44.22\\
        OrdMoE        & 67.00 & 41.44 & 42.29 & 67.90 & 49.82 & 49.30 & 52.96 \\
     \rowcolor{gray!20} Improve($\Delta$) & \textcolor{ForestGreen}{+11.43} & \textcolor{ForestGreen}{+7.88} & \textcolor{ForestGreen}{+11.66} & \textcolor{ForestGreen}{+7.00} & \textcolor{ForestGreen}{+4.32} & \textcolor{ForestGreen}{+10.17} & \textcolor{ForestGreen}{+8.74} \\
        \bottomrule
    \end{tabular}
\end{table}
\section{Conclusion}

We have presented OrdMoE, a novel preference alignment framework that leverages the intrinsic routing dynamics of Mixture-of-Experts (MoE) architectures to enable fully self-supervised alignment of multimodal large language models. By observing that expert selection scores naturally encode a quality-aware hierarchy, OrdMoE constructs an internal ordinal preference ordering over model responses, without requiring human annotations, teacher models, or engineered input perturbations. Our method introduces minimal modifications to the training objective, yet consistently improves performance across a diverse suite of multimodal benchmarks. The results demonstrate that the MoE router, long treated merely as a sparse activation mechanism, in fact harbors rich supervisory signals for alignment. We hope OrdMoE inspires further exploration of intrinsic model dynamics as a scalable and annotation-free path toward aligned, capable multimodal systems.

{
    \small
    \bibliographystyle{ieeenat_fullname}
    \bibliography{main}
}

\clearpage
\setcounter{page}{1}
\maketitlesupplementary

\begin{figure*}[!t]
    \centering
    \includegraphics[width=0.95\linewidth]{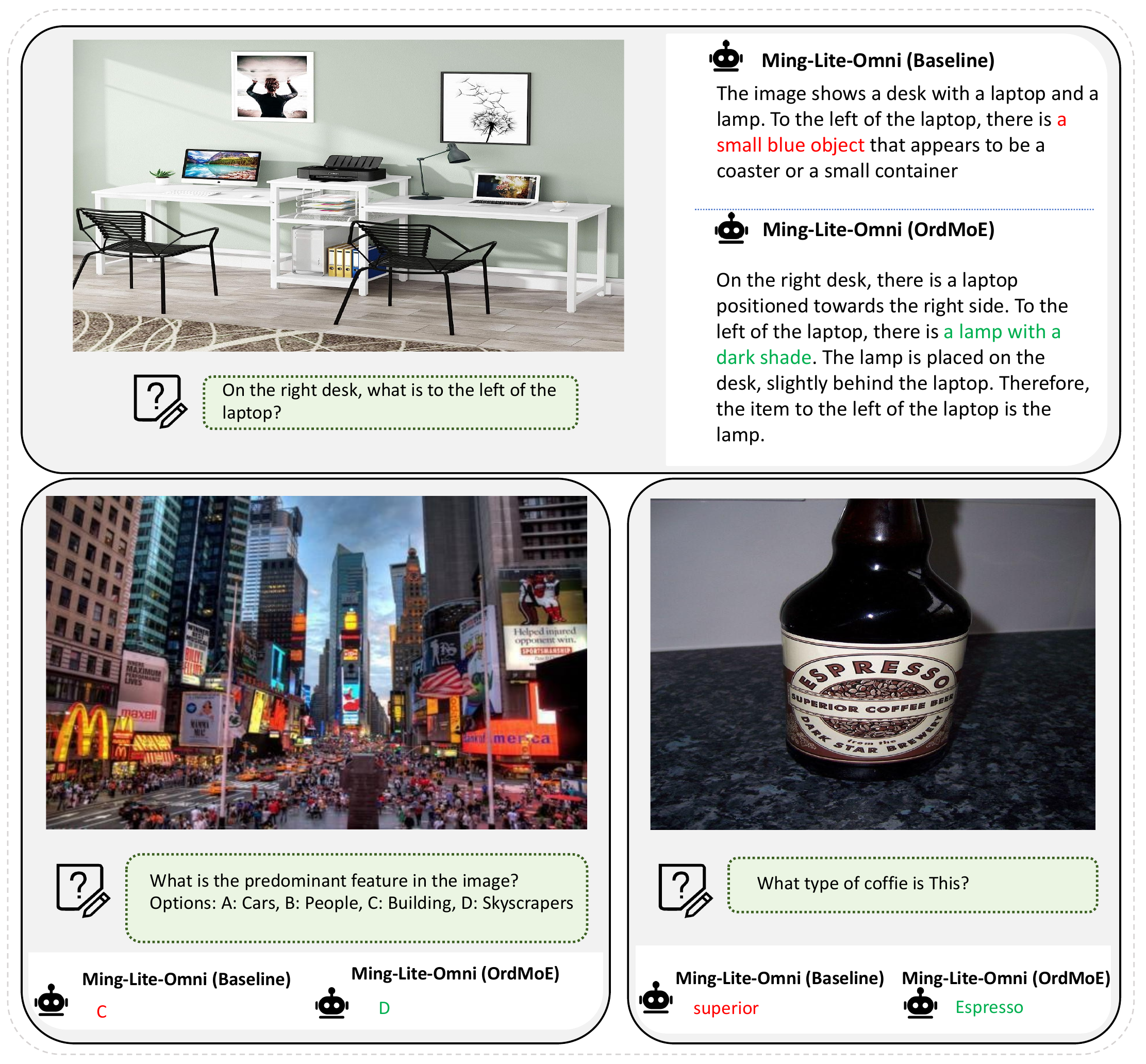}
    \caption{Comparing results between OrdMoE and Baseline. Our method
effectively enhances the general performance of the base model, demonstrating significant improvements in localization precision, fine-grained recognition, and OCR capabilities.}
    \label{fig:preferelearning_demo}
\end{figure*}

\section{Training Details}
\label{sec:training_details}
\subsection{Multimodal Corpus}
\label{subsec:training_dataset}

Section 4.3 of the main text reports experimental results across all modalities. Here, we provide additional details on the data scale underlying our full training pipeline. Specifically, pretraining is conducted on 0.11 trillion multimodal tokens, followed by supervised fine-tuning on a curated instruction-following dataset consisting of 3 million image–text pairs, 1 million video–text pairs, and 23 hours of speech–text data.

\subsection{Training Setup}
\label{subsec:training_setup}
The details of the training hyperparameters used in our training are presented in~\cref{tab:training_details}.

\begin{table}[htbp]
  \centering
  \caption{\textbf{Training hyperparameters used in our experiments.}}
  \label{tab:training_details}
  \renewcommand{\arraystretch}{0.93}
  \resizebox{\columnwidth}{!}{%
    \begin{tabular}{l|cc}
      \toprule
      \diagbox{Setting}{Model} & \multicolumn{2}{c}{\textbf{Ming-Lite-Omni}} \\ 
      \midrule
      LLM & \multicolumn{2}{c}{Ling-lite} \\
      Vision encoder & \multicolumn{2}{c}{Qwen2.5-vl visual backbone} \\
      Projector & \multicolumn{2}{c}{mlp2x\_gelu} \\
      \midrule
      Initial Learning rate & \multicolumn{2}{c}{2e-5} \\
      Batch size per GPU & \multicolumn{2}{c}{16} \\
      Trainable parameters & \multicolumn{2}{c}{all parameters} \\                           
      Learning rate scheduler & \multicolumn{2}{c}{Cosine} \\
      Optimizer & \multicolumn{2}{c}{AdamW} \\
      \bottomrule
    \end{tabular}
}
  \vspace{-10pt}
\end{table}
\subsection{Evaluation Benchmarks}

This study employs a series of widely recognized benchmarks to comprehensively evaluate the model's multimodal capabilities. The benchmarks are categorized as follows:

\begin{itemize}
    \item \textbf{Comprehensive VQA (MMStar, MMVet)}: assessing complex multimodal understanding and holistic task-solving in diverse scenarios. MMStar uses Accuracy while MM-Vet employs GPT-4 based comprehensive scoring.
    \item \textbf{OCR (\textbf{OCRBench})}: evaluating text recognition robustness across various scenarios.
     \item \textbf{Visual Reasoning (MathVista)}: evaluating visual mathematical reasoning capabilities.
    \item \textbf{Visual multi-discipline (AI2D, MMMU)}: evaluating visual understanding across scientific and multi-disciplinary domains.
     \item \textbf{Video Understanding}:
    \begin{itemize}[noitemsep,topsep=2pt]
        \item \textbf{LongVideoBench}: evaluating long-form video understanding with extended temporal dependencies.
        \item \textbf{DiDeMo}: testing text-to-video temporal localization for moment retrieval.
        \item \textbf{AVQA}: assessing audio-visual question answering across visual and auditory modalities.
        \item \textbf{MVBench}: evaluating comprehensive video reasoning across multiple task types.
        \item \textbf{Video-MME}: assessing multi-dimensional capabilities spanning perception and reasoning.
    \end{itemize}
    
    \item \textbf{Speech Recognition (\textbf{Aishell1}, \textbf{LibriSpeech})}: evaluating speech recognition capabilities in both Mandarin Chinese and English. Aishell1 uses character error rate (CER) while LibriSpeech employs word error rate (WER) for performance assessment.
\end{itemize}
\section{Additional Case Studies}
In this section, we present additional case studies to further illustrate the effectiveness of OrdMoE in addressing key weaknesses of the base model and enhancing generalization. To ensure a fair comparison, both the baseline and our OrdMoE use greedy decoding for all results.  In the first example (top row), OrdMoE correctly identifies the desk lamp positioned next to the laptop. In the first case of the second row, OrdMoE demonstrates fine-grained visual understanding by accurately recognizing the structure as a skyscraper. The second case in the same row shows a substantial improvement in OCR capability, with the model precisely identifying the specific coffee variety.

\end{document}